\documentclass[conference]{IEEEtran}
\usepackage{cite}
\usepackage{amsmath,amssymb,amsfonts}
\usepackage{algorithmic}
\usepackage{graphicx}
\usepackage{subcaption}
\usepackage{textcomp}
\usepackage{xcolor}
\usepackage{booktabs}
\usepackage{array}
\usepackage{caption}
\usepackage{multirow}
\usepackage{ragged2e}
\usepackage{mathtools}  
\usepackage{bm}         
\usepackage{url}
\usepackage{hyperref}
\usepackage{siunitx} 
\usepackage[numbers,sort&compress]{natbib}
\def\BibTeX{{\rm B\kern-.05em{\sc i\kern-.025em b}\kern-.08em
    T\kern-.1667em\lower.7ex\hbox{E}\kern-.125emX}}
\begin{document}

\title{Knowledge Graph Embeddings with Representing Relations as Annular Sectors}

\author{\IEEEauthorblockN{1\textsuperscript{st} Yingqi Zeng}
~\\
\and
\IEEEauthorblockN{2\textsuperscript{nd} Huiling Zhu*}
*Corresponding author
~\\
}

\maketitle

\begin{abstract}
Knowledge graphs (KGs), structured as multi-relational data of entities and relations, are vital for tasks like data analysis and recommendation systems. Knowledge graph completion (KGC), or link prediction, addresses incompleteness of KGs by inferring missing triples \textit{((h, r, t)}. It is vital for downstream applications. Region-based embedding models usually embed entities as points and relations as geometric regions to accomplish the task. Despite progress, these models often overlook semantic hierarchies inherent in entities.

To solve this problem, we propose \textbf{SectorE}, a novel embedding model in polar coordinates. Relations are modeled as annular sectors, combining modulus and phase to capture inference patterns and relation attributes. Entities are embedded as points within these sectors, intuitively encoding hierarchical structure. Evaluated on FB15k-237, WN18RR, and YAGO3-10, SectorE achieves competitive performance against various kinds of models, demonstrating strengths in semantic modeling capability.
\end{abstract}

\begin{IEEEkeywords}
component, formatting, style, styling, insert
\end{IEEEkeywords}

\section{Introduction}
Knowledge graphs, as a typical representative of big data knowledge engineering, play a significant role in various tasks such as data analysis, intelligent search, smart recommendations, and intelligent question answering, aligning with the developmental trends of the big data era.

In knowledge graphs, entities serve as graph nodes, indicating real-world objects or abstract concepts; relations act as graph edges, indicating connections between entities, thus forming a multi-relational directed graph composed of various types of entities and relations. The most fundamental form of knowledge storage is the factual triple, denoted as $r(e_h, e_t)$, which signifies that the head entity $e_h$ is connected to the tail entity $e_t$ through the relation $r$.

Due to the limited variety and quantity of data sources, as well as the inaccuracy of algorithms used in the knowledge extraction process from these data sources, knowledge graphs have always faced issues with incomplete structure and content. This incompleteness of information significantly constrains the potential of knowledge graphs in downstream applications. Therefore, knowledge graph completion(KGC), which aims to infer missing facts within the graph based on known facts, also known as link prediction in knowledge graphs, has become a vital area of research.

Several models have been proposed to accomplish the link prediction task. As an effective approach, knowledge graph embedding models map entities and relations to continuous vector spaces. The earliest embedding model, TransE\cite{TransE}, learns a vector for each entity and relation and evaluates the plausibility of a given triple through a scoring function. The vector representations $\textbf{e}, \textbf{r} \in \mathbb{R}^d$ are the embeddings of entity and relation with the same dimension. Inspired by TransE, in order to capture more information about entities and relations as well as the interactions among them, there are rich variations in the embedding methods, yet they still adhere to a similar paradigm.

Later, region-based embedding models\cite{BoxE} are developed, potentially utilizing fuzzy logic to model relations. They represent each relation \textit{r} as a geometric region $X_n \subseteq \mathbb{R}^{2n}$ to naturally model rules in KGs. Regarding the problem of Existential Positive First-order query(EPFO query) in KGs, Query2Box\cite{query2box} encodes the query into a box, where a set of points within the box corresponds to a set of answer entities in the query. Employing a similar approach to accomplish the link prediction task, BoxE\cite{BoxE} encodes relations into boxes and entities into points, with the head entity and tail entity of the relation's corresponding triples being points within the head box and tail box respectively. These embedding models spatially characterize basic logical properties. However, these models focus more on designing the geometric regions that represent relations, and do not delve too deeply into the information of entities. In fact, the embeddings of entities also contain a great deal of information, e.g., semantic hierarchy of entities in a knowledge graph. 

Semantic hierarchy is a potentially important attribute within knowledge graphs, which can be used to assist determining the correctness of the answers in link prediction. Entities inherently possess semantic hierarchy, where some entities have a higher semantic hierarchy than others. For instance, ``animal'' has a higher hierarchy than ``mammal'', and ``mammal'' has a higher hierarchy than ``dog'', thus naturally ``animal'' has a higher semantic hierarchy than ``dog''. Thus, it is important to distinguish the semantic hierarchies of the entities in order to help predicting the missing triples in KGs. Previous studies have proposed several methods to model the semantic hierarchical structure of entities in knowledge graphs. HAKE\cite{HAKE} maps entities and relations into the polar coordinate system. They use a module part to model entities with different semantic hierarchies and a phase part to distinguish entities with the same semantic hierarchy. HRQE\cite{HRQE} maps entities into the 3D polar coordinate system so that the model has the property of hierarchy awareness.

To better obtain semantic hierarchy about entities and their interactions, we embed entities in the polar coordinate system and naturally set the relation regions according to modulus part and phase part. That is, our embeddings of relations form annular sectors while embeddings of entities turn to be points within it. We called our model \textbf{SectorE}, a novel region-based embedding model. In addition to common symmetry/anti-symmetry relation, our model has the ability to capture multiple relation patterns such as subsumption and intersection. Our contributions are as follows:
\begin{itemize}
\item We map the entities into the polar coordinate system to capture their semantic information such as hierarchy and set up annular sector shaped regions for relations to characterize the interactions between them.
\item Our model captures inference patterns with ease through intuitive interactions between annular sectors. Moreover, we naturally express many relation attributes through the properties of relation regions.
\item We evaluate the performance on link prediction task of our model on FB15k-237\cite{fb15k-237}, WN18RR\cite{wn18rr}, and YAGO3-10\cite{yago3-10}, which are the most commonly used benchmarks for knowledge graph completion models. Our model achieved performance against various kinds of models on all these datasets and outperformed them in some metrics.
\end{itemize}

\section{Related Work}
We generalize the knowledge graph completion models into four categories: translational distance models, semantic matching models, neural network models and models based on pre-trained language models.

\subsection{Translational Distance Models}
Translational distance models map the entities and relations in the knowledge graph to continuous vector spaces, defining the relations as the distance from the head entity to the tail entity. The purpose is to make the head entity as close as possible to the tail entity in the vector space after the transition of the relation. TransE is the first model of this kind and has been widely applied later. Although it is simple and efficient, this model still has many limitations, e.g., it composes that the tail entities corresponding to the same relation for a head entity are all equal. TransH\cite{TransH} develops it by modeling a relation as a hyperplane together with a translation operation on it. Based on previous work, TransR\cite{TransR} builds entity and relation embeddings in separate entity space and relation space since an entity may have multiple semantic aspects and thus various triples may focus on different aspects of entities due to different relations. TransD\cite{TransD} improves TransR by constructing dynamic projection matrices for the head and tail entities respectively, enriching the interaction between entities and relations. Furthermore, to capture more relation patterns including symmetry/asymmetry, inversion, and composition, RotatE defines each relation as a rotation from head entities to tail entities in a complex vector space. In addition to these relation patterns, HAKE\cite{HAKE} focuses more on semantic hierarchies. It maps entities into the polar coordinate system such that it can distinguish the hierarchical level of the entities. HRQE\cite{HRQE} maps entities into the 3D polar coordinate system to achieve the property of hierarchy awareness.

Region-based knowledge graph embedding model is a kind of translational distance model that learns a geometric region for each relation. It differs in utilizing the distance the entity points far from the boundaries of the relation region as a distance function. Query2Box\cite{query2box} first sets query as a box and a set of answer entities as points within the box to achieve logical reasoning on KGs. Similar to Query2Box, BoxE\cite{BoxE} embeds entities as points, and relations as a set of boxes to accomplish the link prediction task. However, BoxE is not capable of capturing composition pattern. ExpressivE\cite{ExpressivE} embeds pairs of entities as points and relations as hyper-parallelograms in the virtual triple space $\mathbb{R}^{2d}$. Octagon embeddings models\cite{Octagon} represents relations as regions composed of octagons aligned with the axis, which easily computes intersection and composition. Compared with the previous models, these embedding models better characterize logical rules.

\subsection{Semantic Matching Models}
The main idea of semantic matching models is to determine the likelihood of triples by measuring the compatibility of entities and relations in the semantic space. Such models quantify the semantic matching among the entities and relations so as to predict the missing links. 
RESCAL\cite{RESCAL} is a typical semantic matching model representing each relation as a full-rank matrix and obtains the score function by matrix multiplication. Based on this approach, DistMult\cite{DistMult} reduces the number of parameters of RESCAL by restricting the relation matrix to be a diagonal matrix. The downside that follows is that all relations are restricted to be symmetric. To further address the issue of symmetric relations, ComplEx\cite{Complex} generalizes the approach of DistMult to the complex vector space. The real part of embedding is symmetric, while the imaginary part is asymmetric. Therefore, it can model both symmetric and asymmetric relations. SimplE\cite{SimplE} improves the tensor decomposition by learning two embeddings for each entity dependently, and proves itself to be fully expressive. DualE\cite{DualE} introduces hypercomplex representations to model entities and relations to enhance expressiveness of the model. 

\subsection{Neural Network Models}
Although traditional link prediction models are simple in calculation and highly interpretable, they have difficulty in capturing more complex interactions, and there is limited room for improvement in prediction accuracy. In order to achieve better performance of the task, many studies have applied neural networks to link prediction, including traditional neural network models and models based on pre-trained language models. 

\subsubsection{Traditional Neural Network Models}
ConvE\cite{ConvE} is the first to apply convolutional layers to KGC. ConvE utilizes 2D convolution operation to increases the expressiveness of the model through additional points of interaction between embeddings. ConvKB\cite{ConvKB} uses 1D convolution instead to extract features of entities. R-GCN\cite{R-GCN} is a generalization of graph convolutional network. When modeling relational data, R-GCN takes into account the different impacts of the proximity of relations on information propagation.
RSN\cite{RSN} integrate recurrent neural networks with residual learning to capture long-term relational dependencies between entities and relations.

\subsubsection{Models Based on Pre-trained Language Models}
With the advancement of deep learning, numerous researches integrate pre-trained language models such as BERT\cite{BERT} to introduce large-scale external corpus information and achieve significant improvements in link prediction task. KG-BERT\cite{KG-BERT} fine-tunes BERT for KGC and turns KGC into a sequence classification problem. Specifically, it takes entities and relations as textual inputs and computes the plausibility of the triples. In contrast to the previous models that concentrated on textual information, StAR\cite{StAR} augments textual encoding paradigm with graph embedding technique while utilizing spatial transfer approach of translational distance models to capture the distances between triples. In addition, there are also methods that do not require training and only use Large Language Models (LLMs). KICGPT\cite{KICGPT} proposes a framework that integrates a LLM and a triple-based KGC retriever. By encoding part of the KG into demonstrations, KICGPT designs an in-context learning strategy called Knowledge Prompt for KGC.

\section{Method}

\subsection{Notations}
We first define a knowledge graph $\bm{G} = \{ \bm{E}, \bm{R}, \bm{T} \}$, where $\bm{E}$ represents the set of entities, $\bm{R}$ the set of relations, and $\bm{T}$ the set of triples within this KG. A triple is in the form of $r(e_h, e_t)$.

\subsection{Region Embeddings}
Inspired by HAKE\cite{HAKE}, we embed entities into the polar coordinate system to capture the semantic hierarchies of entities. In correspondence with the embeddings of entities, we model each relation as a geometric region, and the shape of the region is an annular sector in polar coordinates. In a knowledge graph, the number of entities is much larger than that of relations. One relation corresponds to a large number of triples and numerous different entities. Therefore, compared with modeling a relation as a single point in the vector space, a region may have a higher tolerance for entities at different semantic levels and is more conducive to capture different triples.

\subsubsection{Entity Representation}\label{sec:ent.embed}
In a knowledge triple, the position of each entity is influenced by itself and the transformation of the other entity in the triple. Thus, every entity $e_i \in \mathbf{E}$ is represented by two vectors: a basic vector $ \bm{e_i}$ representing the initial position of the entity and a transformational bump $\bm{b_i}$ representing the transformational effect to other entities. Since entities are mapped into the polar coordinate system, $ \bm{e_i} = (\bm{m_{e_i}}, \bm{p_{e_i}}) \in \mathbb{R}_+^d \times [0, 2\pi)^d$ and $\bm{b_i}=(\bm{m_{b_i}}, \bm{p_{b_i}}) \in \mathbb{R}_+^d \times [0, 2\pi)^d $.
Consequently, the ultimate embedding of the entity $e_i$ in a certain triple $r(e_i, e_j)$ is defined as:

\begin{equation}
e_i^{r(e_i,e_j)} = e_i \odot b_j
\end{equation}

To be exact, the specific embedding $e_i^{r{(e_i,e_j)}}$ can be represented as follows:

\begin{equation}
\begin{aligned}
    \bm{e_i^{r(e_i,e_j)}} &= (\bm{m_{e_i}} \odot \bm{m_{b_j}}, \quad (\bm{p_{e_i}} + \bm{p_{b_j}})\mod 2\pi)
\end{aligned}
\end{equation}

Note that the ultimate embedding $e_i^{r{(e_i,e_j)}}$ depends on not only the basic vector $\bm{e_i}$, but also the transformational bump $\bm{b_j}$ from entity $e_j$. Thus, the entity embedding varies according to different triples.

\subsubsection{Relation Representation}
According to the way we embed entities, we embed relations as annular sectors. Specifically, every relation $r \in \bm{R}$ is defined by two annular sectors: the head sector $\bm{r^h}$ and the tail sector $\bm{r^t}$.

An annular sector $\bm{r}$ is determined by the boundaries of modulus and phase. The upper and lower boundaries of modulus are denoted as $\bm{u(r)},\bm{l(r)}$. The central modulus is \( \bm{c(r)} = \left( \bm{u(r)} + \bm{l(r)} \right) / 2 \) and the annular depth is \( \bm{w(r)} = \bm{u(r)} - \bm{l(r)} + 1 \). The upper and lower boundaries of phase are denoted as $\bm{\phi(r)}, \bm{\psi(r)}$ respectively. The central phase is \( \bm{\theta(r)} = \left( \bm{\phi(r)} + \bm{\psi(r)} \right) / 2 \) and the central angle of the annular sector is \( \bm{\delta(r)} = \bm{\phi(r)} - \bm{\psi(r)} + \beta\pi \), where $\beta$ is a hyperparameter.

\begin{figure}
\centering
\includegraphics[width=0.6\columnwidth]{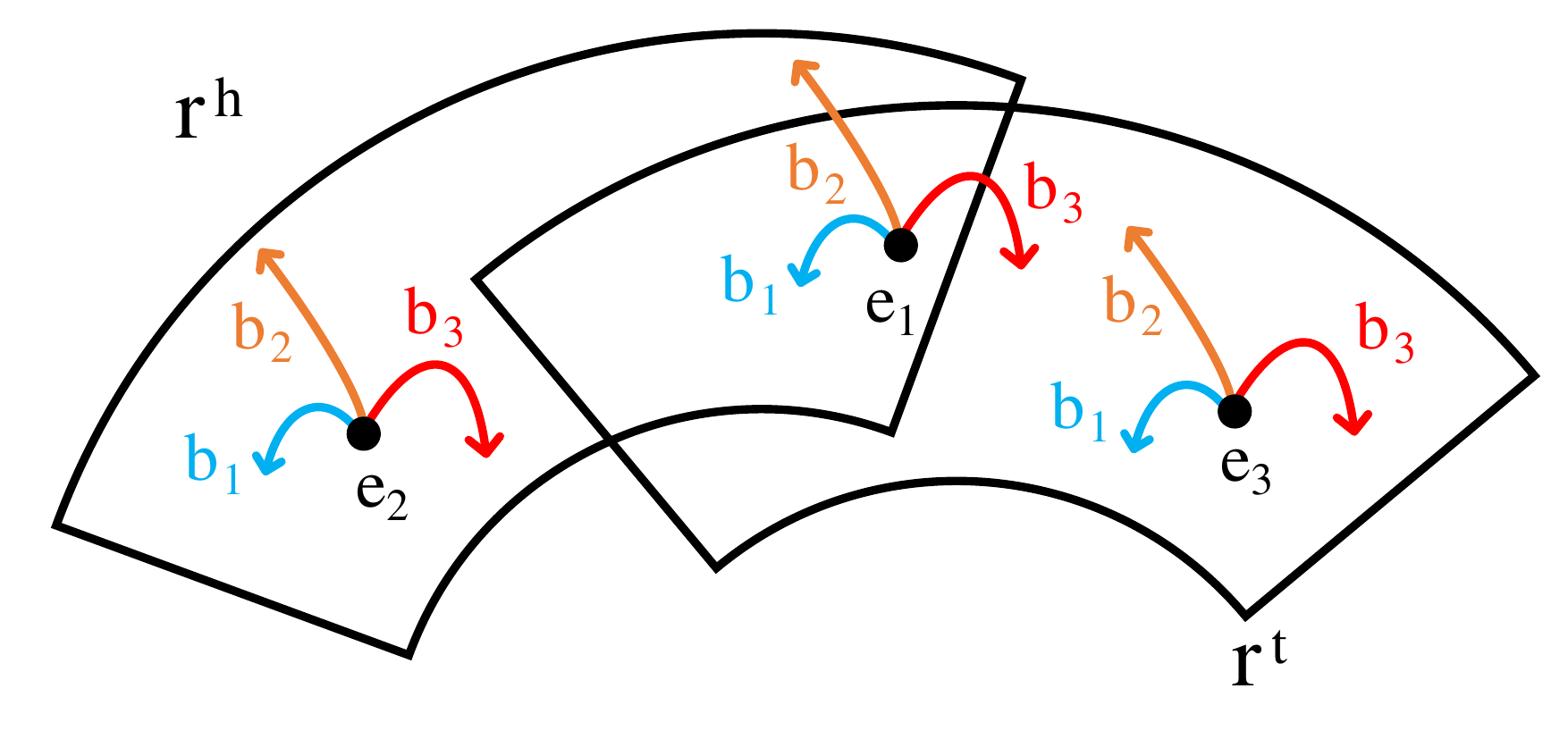}
\caption{Simple illustration of SectorE. $\bm{r^h}$ and $\bm{r^t}$ are the region embeddings of the relation $r$. The position of the entity $e_i$ is determined by a basic point $\bm{e_i}$ and a transformational bump $\bm{b_j}$. }
\label{fig:model}
\end{figure}

As shown in Fig.~\ref{fig:model}, we give a concrete example over a relation $r$ and entities $e_1$, $e_2$ and $e_3$ to illustrate our model. If the head entity point after bumping from the tail entity lies within the head sector $\bm{r^h}$ and the tail entity point after bumping from the head entity lies within the tail sector $\bm{r^t}$, the triple holds. For instance, $r(e_2, e_3)$ is satisfied in our model, since both $e_2^{r{(e_2,e_3)}} = e_2 \odot b_3$ is a point in $\bm{r^h}$ and $e_3^{r{(e_2,e_3)}} = e_3 \odot b_2$ is a point in $\bm{r^t}$. Similarly, after bumping with $b_1$, $e_1$ landed in the overlapping area of the two annular sectors. Thus, we can say that $r(e_1, e_1)$ is a true fact. By contrast, $r(e_2, e_1)$ is false since $e_1^{r{(e_2,e_1)}}$ is not in $\bm{r^h}$. 

\subsubsection{Scoring Function}
The goal of our model is to make the head entity and the tail entity of the correct triple fall within the corresponding annular sectors as much as possible, and make the entities of the incorrect triples stay as far away as possible from the relation regions. Therefore, similar to BoxE\cite{BoxE}, we use the distance from the entity points to the center of the region to measure the score of the predicted triples. However, we calculate the distances of the modulus and phase parts separately.

In the modulus part, we aim to model entities at different hierarchical levels using the annular depth of their coordinates. The depth-dependent factor $\bm{w(r)}$ is used to make entity points move rapidly into their target regions. After reaching the target regions, the entity points are supposed to hold the positions. The distance function for the modulus part is:

\begin{equation}
\mathrm{dist}_{m}\left(\bm{e_{h}^{r(h,t)}}, \bm{r^h} \right) = 
\begin{cases}
\begin{aligned}
&\left| \bm{m_{e_h}} - \bm{c(r^h)} \right| \oslash {\bm{w(r^h)}}, \\
&\qquad\qquad\qquad\qquad \text{if } \bm{e_{h}^{r(h,t)}} \in \bm{r^h} \\[1.5ex]
&\left| \bm{m_{e_h}} - \bm{c(r^h)} \right| \odot \bm{w(r^h)} - \bm{k}, \\ &\qquad \text{otherwise}
\end{aligned}
\end{cases}
\end{equation}
where $\bm{k} = 0.5 (\bm{w(r^h) }- \bm{1}) \odot (\bm{w(r^h)} - \bm{1} \oslash \bm{{w(r^h)}}) $, $\bm{1}$ represents the vector with every place $1$. We use $k$ to ensure the continuity of the piecewise distance function at the boundary points. $\mathrm{dist}_{m}\left(\bm{e_{t}^{r(h,t)}}, \bm{r^t} \right)$ is defined similarly.

In terms of phase part, we use a sine function to measure the distance between phases to unify the scoring criteria of modulus and phase, following the same evaluation method with pRotatE\cite{RotatE}. The distance function for the phase part is:

\begin{equation}
\mathrm{dist}_p\left(\bm{e_h^{r(h,t)}}, \bm{r^h}\right) = 
\begin{cases}
\begin{aligned}
&\left| \ \sin\left( \frac{\bm{p_{e_h}} - \bm{\theta(r^h)}}{2} \right) \right| \ \oslash{\bm{\delta(r^h)}}, \\
&\qquad\qquad\qquad\qquad \text{if } \bm{e_{h}^{r(h,t)}} \in \bm{r^h} \\[1.5ex]
&\left| \sin\left( \frac{\bm{p_{e_h}} - \bm{\theta(r^h)}}{2} \right) \right| \odot \bm{\delta(r^h)} - \bm{k},\\
&\qquad\qquad\qquad\qquad \text{otherwise}
\end{aligned}
\end{cases}
\end{equation}
where $\bm{k} = 0.5 (\bm{\delta(r^h)} - \bm{1} \oslash \bm{\delta(r^h)})$. $\mathrm{dist}_{p}\left(\bm{e_{t}^{r(h,t)}}, \bm{r^t} \right)$ is defined similarly.

By combining the modulus part and the phase part, the total scoring function of is:

\begin{equation}
\begin{multlined}
\mathrm{score}(r(h,t)) = - \sum_{i=h,t} ( \lambda_1\ \left\| \mathrm{dist}_{m} \left( \bm{e_i^{r(h,t)}}, \bm{r^i} \right) \right\| \\
+ \lambda_2 \, \left\|\mathrm{dist}_{p} \left( \bm{e_i^{r(h,t)}}, \bm{r^i}\right) 
 \right\| )
\end{multlined}
\label{eq:score}
\end{equation}
$\lambda_1, \lambda_2 \in \mathbb{R}$ are parameters learned by the model.

\subsection{Loss Function}
To train the model, we use the negative sampling loss functions with self-adversarial training\cite{RotatE}:

\begin{equation}
\begin{aligned}
L = {} & -\log \sigma \left( \gamma + \mathrm{score}(r(h, t)) \right) \\
& - \sum_{i=1}^n p(h_i', r, t_i') \log \sigma \left( -\mathrm{score} \left( r(h_i', t_i')-\gamma \right) \right)
\end{aligned}
\label{eq:combined}
\end{equation}
where \(\gamma\) is a fixed margin, \(\sigma\) is the sigmoid function, and \((h'_i, r, t'_i)\) is the \(i\)-th negative triple. Specifically, 

\begin{equation}
p\left( h_{j}^{\prime}, r, t_{j}^{\prime} \mid \{ h_{i}, r_{i}, t_{i} \} \right) = 
\frac{
  \alpha \left( \operatorname{score} \left( r \left( h_{j}^{\prime}, t_{j}^{\prime} \right) \right) \right)
}{
  \sum_{i} \exp \left( \alpha \left( \operatorname{score} \left( r \left( h_{i}^{\prime}, t_{i}^{\prime} \right) \right) \right) \right)
}
\label{eq:probability}
\end{equation}
is the probability distribution of sampling negative triples, where $\alpha$ is the temperature of sampling.

\subsection{Model Capabilities}
The expressive ability of a model is usually demonstrated by its capability to capture relation patterns. We will introduce the definition of some relation patterns and prove that our model is able to capture them through interactions between regions.

Let $\bm{r_i}$ define the relation region of the relation \( r_i \in \textbf{R} \), $\bm{r_i^h}$ the head sector, $\bm{r_i^t}$ the tail sector. As mentioned before, $r_i(x, y)$ holds \emph{iff} 
$\bm{e_x^{r_i(x,y)}} \in \bm{r_i^h}$ and $\bm{e_y^{r_i(x,y)}} \in \bm{r_i^t}$.

\subsubsection{Symmetry}
$r_1(x,y) \to r_1(y,x)$, e.g., ``is similar to'' and ``has neighbor''. Our model captures symmetry relation with two identical regions: $\bm{r_1^h}$ = $\bm{r_1^t}$.

\subsubsection{Anti-symmetry}
$r_1(x,y) \to  \lnot r_1(y,x)$, e.g., ``is parent of'', ``is child of''. Our model captures anti-symmetry relation by setting: \( \bm{r_1^h} \cap \bm{r_1^t} = \emptyset \).

\subsubsection{Inversion}
$r_1(x,y) \leftrightarrow r_2(y,x)$. An example of a pair of inversion relations:``is parent of'' and ``is child of''. Our model achieves it by letting: \( (\bm{r_1^h} = \bm{r_2^t}) \land (\bm{r_1^t} = \bm{r_2^h}) \).

\subsubsection{Subsumption}
$r_1(x,y) \to r_2(x,y)$ (also known as hierarchy). An example of subsumption relation pair is: ``is parent of'' and ``is elder of''. Our model captures it by setting: \( (\bm{r_1^h} \subseteq \bm{r_2^h}) \land (\bm{r_1^t} \subseteq \bm{r_2^t}) \).

\subsubsection{Intersection}
$r_1(x,y) \land r_2(x,y) \to r_3(x,y)$. A set of examples of $r_1, r_2,r_3$ is:``was born on the same day'', ``has same parents with'' and ``is a twin with''. Our model captures it by setting: \( ((\bm{r_1^h} \cap \bm{r_2^h}) \subseteq \bm{r_3^h}) \land ((\bm{r_1^h} \cap \bm{r_2^h}) \subseteq \bm{r_3^h}) \).

\subsubsection{Mutual exclusion}
$r_1(x,y) \land r_2(x,y) \to \bot$. An example of mutual exclusion relation pair is: ``is father of'' and ``is mother of''. Our model captures it through disjoint regions: \( (\bm{r_1^h} \cap \bm{r_2^h} = \emptyset) \lor (\bm{r_1^t} \cap \bm{r_2^t} = \emptyset) \).

\section{Experiments}

\subsection{Experimental Settings}
\subsubsection{Datasets}
We conduct experiments of our model on three widely-used knowledge graph completion benchmarks: FB15k-237, WN18RR and YAGO3-10. The basic information and the split on these datasets are listed in Table\ref{tab:datasets}. 
FB15k-237 is a subset of FB15k\cite{TransE} with real-world commonsense knowledge. WN18RR is a subset of WN18\cite{TransE}, a lexical database.
YAGO3-10 is a subset of YAGO3\cite{yago3-10} with a large number of personal attributes. Among the benchmark datadsets, FB15k-237 and WN18RR are preprocessed by removing inverse relations to prevent the data leakage problem.

\begin{table}[ht]
\centering
\caption{Information of Used Benchmark Datasets}
\label{tab:datasets}
\begin{tabular}{@{} l *{5}{r} @{}}
\toprule
Dataset    & \#Ent   & \#Rel   & \#Train  & \#Valid & \#Test \\
\midrule
FB15K-237  & 14,541  & 237     & 272,115  & 17,535  & 20,466  \\
WN18RR     & 40,943  & 11      & 86,835   & 3,034   & 3,134   \\
YAGO3-10   & 123,182  & 37      & 1,079,040  & 5,000   & 5,000   \\
\bottomrule
\end{tabular}
\end{table}

\subsubsection{Evaluation protocol}
We evaluate models under the ``Filter'' setting, where known true triples are excluded from rankings. Given a test triplet $r_i(e_h, e_t)$, we generate candidate predictions by substituting either the head or tail position with the entities set $\bm{E}$, resulting in two candidate sets: head prediction queries: \( \{r_i(e, e_t) | e \in \bm{E}\} \), and tail prediction queries: \( \{r_i(e_h, e) | e \in \bm{E}\} \). Each candidate triplet receives a compatibility score through the model's scoring function. Two metrics are considered: (1)MRR (Mean Reciprocal Rank): average reciprocal rank of correct entities, (2)Hits at N (Hits@N, N=1, 3, 10): proportion of correct entities ranked in top-N predictions. The larger MRR and Hits@N, the more accurately the model predicts.

\subsubsection{Training Setting}
We follow the implementation details of RotatE\cite{RotatE}, using Adam optimizer and grid search to find the best hyperparameters. The ranges of the hyperparameters for the grid search are set as follows: The batch size of each epoch is selected in \{256, 512,1024\}, the embedding dimensions is chosen from \{50, 100, 200, 500, 1000\}, the number of negative samples for every positive sample is selected in \{256, 512, 1024\}, fixed margin $\gamma$ is chosen from \{3, 6, 9, 12, 18, 24, 30\}, self-adversarial sampling temperature $\alpha$ is selected in \{0.5, 1\}.

\subsubsection{Baselines}
We compare our model with several strong baselines. For translational distance models, we report TransE\cite{TransE}, RotatE\cite{RotatE}, HAKE\cite{HAKE} and BoxE\cite{BoxE}. For semantic matching models, we report Dismult\cite{DistMult} and ComplEx\cite{Complex}.

\subsection{Main Results}
We report experimental results of all baseline models and our model on FB15k-237, WN18RR and YAGO3-10 in Table \ref{tab:results}. The best scores are indicated in bold, and the second-best scores are underlined. The scores of all baselines are obtained from their corresponding paper. As a region-based model, our model achieves competitive or superior performance when compared with translational distance models and semantic matching models. Table \ref{tab:results} illustrates that our model surpasses previous baseline models in some metrics, showcasing the modeling capabilities of the proposed model. 

YAGO3-10 datasets contains most triples among these datasets and includes various relation patterns. In this dataset, the entity sets corresponding to many relations are extremely large, which is suitable for region-based models. Hence, we can expect that our model is capable of working well on this dataset and results show that it outperforms all baselines in H@1 and MRR as expected. Note that among the baselines, the region-based model BoxE also performs well. It also proves that region-based models are promising in this kind of database.

On the FB15k-237 dataset, we can only gain the highest H@3. Compared with WN18RR and YAGO3-10, FB15k-237 dataset has more complex relation types and fewer entities. In this dataset, there are some triples that do not lead to hierarchy transformation.

\begin{table*}[ht]
\centering
\caption{Evaluation results on WN18RR, FB15k-237 and YAGO3-10 datasets.}
\label{tab:results}
\begin{tabular}{@{}l *{3}{rrrr} @{}}
\toprule
 & \multicolumn{4}{c}{WN18RR} & \multicolumn{4}{c}{FB15k-237} & \multicolumn{4}{c}{YAGO3-10} \\
\cmidrule(lr){2-5} \cmidrule(lr){6-9} \cmidrule(l){10-13}
Method & MRR & H@1 & H@3 & H@10 & MRR & H@1 & H@3 & H@10 & MRR & H@1 & H@3 & H@10 \\
\midrule
TransE    & .226 & -- & .501 & .294 & -- & .465 & -- & -- & -- & -- & -- & -- \\
DistMult  & .43  & .39 & .44  & .49  & .241 & .155 & .263 & .419 & .34  & .24  & .38  & .54  \\
ConvE     & .43  & .40 & .44  & .52  & .325 & .237 & .356 & .501 & .44  & .35  & .49  & .62  \\
ComplEx   & .44  & .41 & .46  & .51  & .247 & .158 & .275 & .428 & .36  & .26  & .40  & .55  \\
RotatE    & .476 & .428 & .492 & .571 & .338 & .241 & .375 & .533 & .495 & .402 & .550 & .670 \\
BoxE      & .451 & .400 & .472 & .541 & .337 & .238 & .374 & .538 & .567 & .494 & \textbf{.611} & \textbf{.699} \\
ModE      & .472 & .427 & .486 & .564 & .341 & .244 & .380 & .534 & .510 & .421 & .562 & .660 \\
HAKE      & \textbf{.497} & \textbf{.452} & \textbf{.516} & .582 & \textbf{.346} & \textbf{.250} & .381 & \textbf{.542} & .545 & .462 & .596 & .694 \\
\midrule
Ours      & .475 & .421 & .478 & \textbf{.586} & .342 & .242 & \textbf{.383} & .540 & \textbf{.570} & \textbf{.498} & .600 & .686 \\
\bottomrule
\end{tabular}

\vspace{2mm}
\end{table*}

\subsection{Ablation Studies}
To validate the contribution of each component in our model, we seperately ablate three core elements: modulus part, phase part and transformational bump. Table~\ref{tab:ablation} compares their impacts on FB15k-237, WN18RR and YAGO3-10 under identical hyperparameters.

\begin{table*}[ht]
\caption{Performance impact of removing modulus part, phase part and transformational bump under the same setting in main experiments. Best results per metric highlighted in bold.}
\label{tab:ablation}
\centering
\begin{tabular}{ccc *{4}{S} *{4}{S} *{4}{S}}
\toprule
\multicolumn{1}{c}{\multirow{2}{*}{\textbf{m}}} & 
\multicolumn{1}{c}{\multirow{2}{*}{\textbf{p}}} & 
\multicolumn{1}{c}{\multirow{2}{*}{\textbf{b}}} & 
\multicolumn{4}{c}{\textbf{WN18RR}} & 
\multicolumn{4}{c}{\textbf{FB15k-237}} & 
\multicolumn{4}{c}{\textbf{YAGO3-10}} \\
\cmidrule(lr){4-7} \cmidrule(lr){8-11} \cmidrule(lr){12-15}
& & & {MRR} & {H@1} & {H@3} & {H@10} & {MRR} & {H@1} & {H@3} & {H@10} & {MRR} & {H@1} & {H@3} & {H@10} \\
\midrule
$\checkmark$ & $\checkmark$ &    & 0.410 & 0.365& 0.423 & 0.498 & 0.286 & 0.191 & 0.293 & 0.487 & 0.480 & 0.383 & 0.532 & 0.635 \\

$\checkmark$ &    & $\checkmark$ & 0.446 & 0.408 & 0.441 & 0.537 & 0.308 & 0.196 & 0.342 & 0.501 & 0.506 & 0.402 & 0.545 & 0.627 \\
             
             & $\checkmark$ & $\checkmark$ & 0.453 & \textbf{0.425} & 0.459 & 0.544 & 0.325 & 0.210 & 0.353 & 0.503 & 0.526 & 0.421 & 0.565 & 0.664 \\

$\checkmark$ & $\checkmark$ & \textbf{$\checkmark$} & \textbf{0.475} & 0.421 & \textbf{0.478} & \textbf{0.586} & \textbf{0.342} & \textbf{0.242} & \textbf{0.383} & \textbf{0.540} & \textbf{0.570} & \textbf{0.498} & \textbf{0.600} & \textbf{0.694} \\
\bottomrule
\end{tabular}
\end{table*}

Obviously, transformational bump's absence strongly weaken the performance of model in all metrics and datasets, proving that it is essential to our model. Compared to modulus part, phase part improves the model better especially in WN18RR dataset. 

\subsection{Model Analysis}
\subsubsection{Relation Analysis}
In this section, we will demonstrate some properties of the relation embeddings through the areas of relation regions after training. As shown in Table \ref{tab:area}, we take YAGO3-10 as an example to illustrate. These areas of the relation sectors are computed through the boundaries embeddings of the sectors learned by the model with the results of Table\ref{tab:results}.

The area of a relation depends on not only popularity but also entity hierarchy related to the relation. In most cases, the more entities a relation connects, the larger the region representing it. For example, the more popular relation \textit{wasBornIn} with above 44,000 triples in training datasets has a mean area 1.48 while \textit{hasNeighbor} with about 500 triples possesses a mean area 0.5. However, the area of the relation region does not fully follow this rule, as it is also influenced by entity hierarchy simultaneously. For instance, the relation \textit{hasGender}, although appearing over 60,000 times, solely has mean region area 0.65, while \textit{hasMusicalRole} with about 7,700 facts has sector areas 1.33 and 0.86. In terms of \textit{hasGender}, tail entities are only male/female, which are quite similar semantically. Conversely, \textit{hasMusicalRole} means to represent the roles or instruments that artists good at within music, so it connects tail entities including musical career types and instruments but its head entities are only musicians. For example, tail entities include synthesist, percussion and glockenspiel(a kind of percussion), which imply different semantic hierarchies of entities. As a result, \textit{hasMusicalRole} has a larger tail sector than \textit{hasGender}.

Besides, the ratio of the two relation sectors, $\bm{r^h}$ and $\bm{r^t}$, reveals that the relation $r$ as an $M$-$N$ relation. In detail, $M$-$N$ < 1 tends to denote 1-N relations, e.g., \textit{hasChild} (0.42 vs 0.75) and \textit{hasCurrency} (0.52 vs 0.90). Similarly, $M$-$N$ > 1 tends to represent $N$-1 relations, such as \textit{diedIn} (1.09 vs 0.66). When $M$-$N$ is approximately equal to 1, it may represents $N$-$N$ or 1-1 relations, e.g., \textit{hasCapital} (1.07 vs 1.12).

These results above prove that our model is capable of inferring some characteristics of entities and relations, such as entities' semantic hierarchies and relation attributes.


\begin{figure*}[t]
  \centering
  \begin{subfigure}{0.48\textwidth}
    \includegraphics[width=\linewidth]{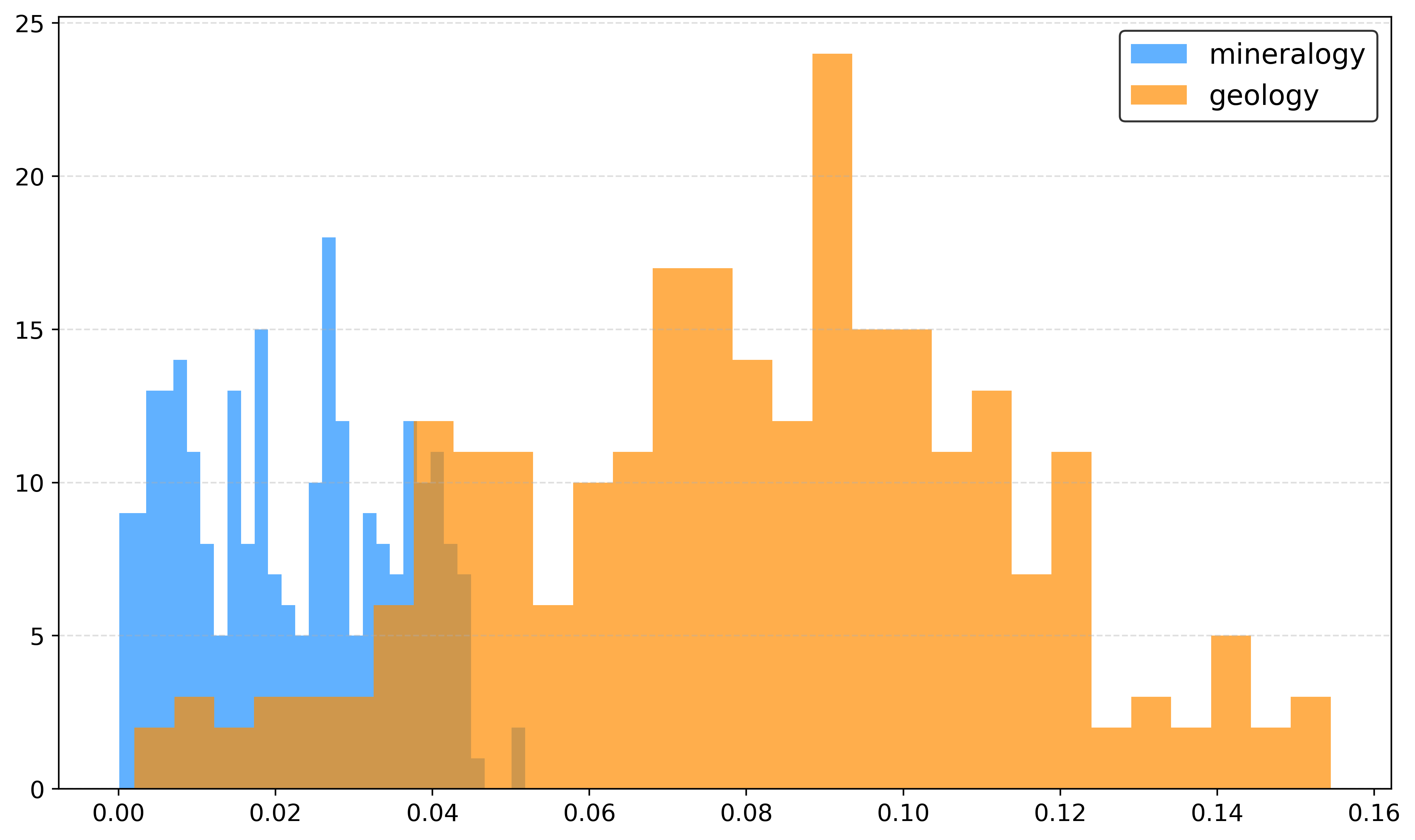}
    \caption{\textit{(mineralogy, hypernym, geology)}}
    \label{fig:sub_a}
  \end{subfigure}
  \hfill
  \begin{subfigure}{0.48\textwidth}
    \includegraphics[width=\linewidth]{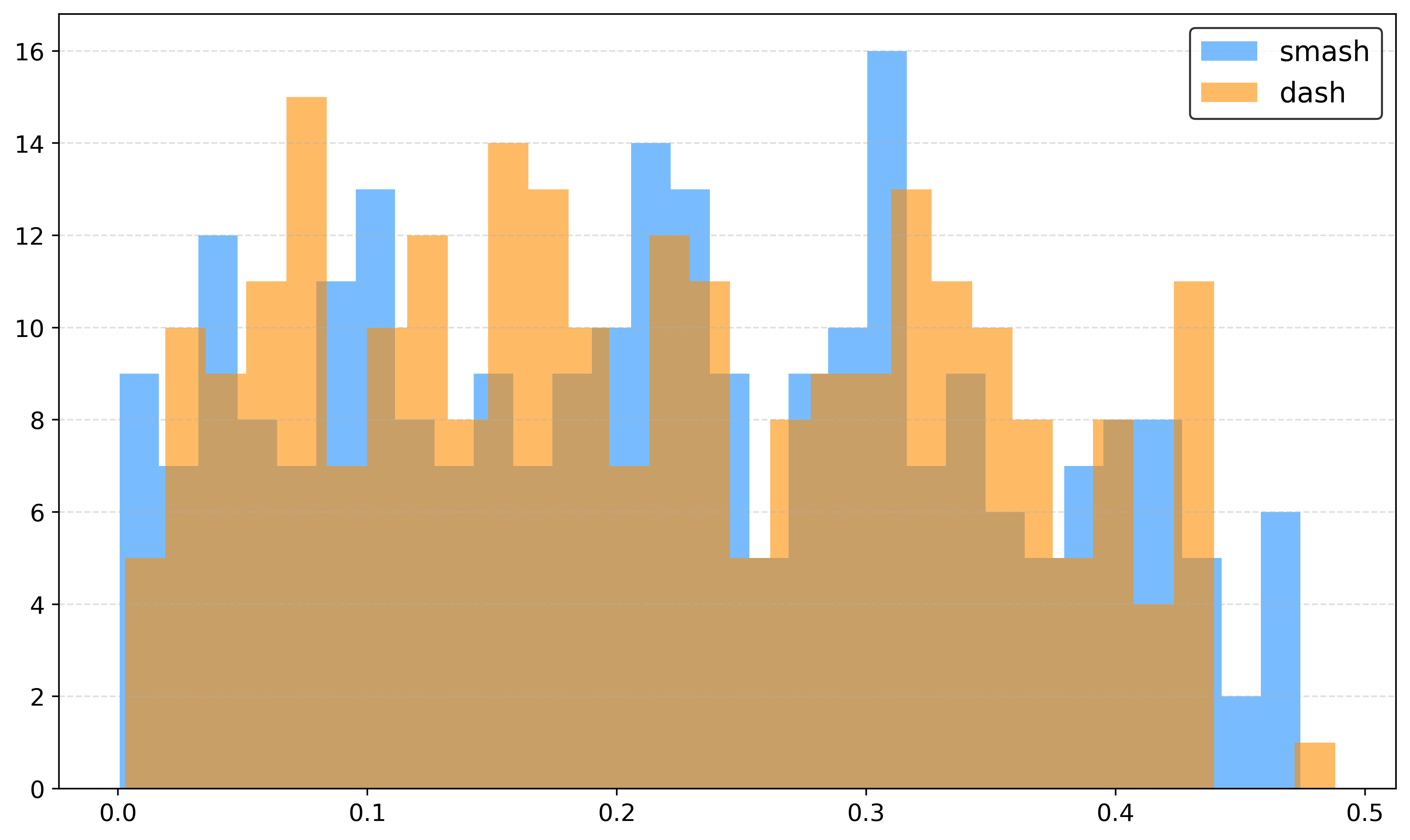}
    \caption{\textit{(smash, verb group, dash)}}
    \label{fig:sub_b}
  \end{subfigure}
  \caption{Visualization of several entity embeddings from WN18RR dataset.}
  \label{fig:main}
\end{figure*}

\begin{table}[t]
\centering
\caption{Geometric mean area per dimension for all relation anunlar sectors in YAGO3-10 after training.}
\label{tab:area}
\footnotesize
\begin{tabular}{@{}>{\RaggedRight}p{2.2cm}cc@{}}
\toprule
\multirow{2}{*}{Relation} & \multicolumn{2}{c}{Box Volume} \\
\cmidrule{2-3}
 & Head Box & Tail Box \\
\midrule
actedIn & 1.3432 & 0.4743 \\
created & 1.1743 & 0.6641 \\
dealsWith & 0.9370 & 0.8107 \\
diedIn & 1.0874 & 0.7773 \\
directed & 0.7739 & 1.1433 \\
edited & 1.1610 & 0.5328 \\
exports & 1.3093 & 0.9782 \\
graduatedFrom & 0.4432 & 0.4547 \\
happenedIn & 0.9296 & 1.1029 \\
hasAcademicAdvisor & 0.7724 & 1.0758 \\
hasCapital & 1.0718 & 1.1236 \\
hasChild & 0.4174 & 0.7505 \\
hasCurrency & 0.5237 & 0.8963 \\
hasGender & 0.8194 & 0.4816 \\
hasMusicalRole & 1.3269 & 0.8623 \\
hasNeighbor & 0.4627 & 0.5513 \\
hasOfficialLanguage & 0.6183 & 1.2992 \\
hasWebsite & 0.3830 & 0.8346 \\
hasWonPrize & 1.2383 & 0.7853 \\
imports & 1.1649 & 0.9389 \\
influences & 0.4690 & 0.7957 \\
isAffiliatedTo & 1.1578 & 1.2348 \\
isCitizenOf & 0.9089 & 1.1650 \\
isConnectedTo & 0.9915 & 0.7586 \\
isInterestedIn & 0.7681 & 0.6348 \\
isKnownFor & 1.1774 & 0.3941 \\
isLeaderOf & 1.2366 & 0.8492 \\
isLocatedIn & 0.4360 & 1.2924 \\
isMarriedTo & 0.5583 & 1.0841 \\
isPoliticianOf & 0.5611 & 1.2513 \\
livesIn & 1.1308 & 0.5074 \\
owns & 0.4354 & 0.8147 \\
participatedIn & 1.0553 & 0.5367 \\
playsFor & 1.0495 & 0.8757 \\
wasBornIn & 1.1601 & 1.3437 \\
worksAt & 1.0253 & 0.6735 \\
wroteMusicFor & 1.3160 & 0.6876 \\
\bottomrule
\end{tabular}
\end{table}

\section{Conclusion}
In this paper, we propose a region-based knowledge graph embedding method SectorE to model multiple relation patterns and semantic hierarchies in knowledge graphs simultaneously. To model the semantic hierarchies in knowledge graphs, we map entities into the polar coordinate system. To model multiple relation patterns in knowledge graphs, we embed relations as annular sectors. Experiments show that our model achieves competitive results on benchmark datasets for the link prediction task. In the future, we intend to explore the performance of our model on further datasets, especially the large ones that including more complex relations.

\end{document}